\documentclass[conference]{IEEEtran}
\IEEEoverridecommandlockouts
\usepackage{cite}
\usepackage{booktabs}
\usepackage[flushleft]{threeparttable}
\usepackage{amsmath,amssymb,amsfonts}
\usepackage{algorithmic}
\usepackage{graphicx}
\usepackage{textcomp}
\usepackage{xcolor}
\usepackage{multirow}
\usepackage{float}
\usepackage[caption=false]{subfig}
\def\BibTeX{{\rm B\kern-.05em{\sc i\kern-.025em b}\kern-.08em
    T\kern-.1667em\lower.7ex\hbox{E}\kern-.125emX}}
\begin{document}

\title{Unsupervised Pre-Training Using Masked Autoencoders for ECG Analysis 
%\title{An Unsupervised Pre-Training Technique Based on Masked Autoencoders for ECG Analysis 
}

\author{Guoxin Wang, \IEEEmembership{Member, IEEE,}
        Qingyuan Wang, \IEEEmembership{Member, IEEE,} Ganesh Neelakanta Iyer, \IEEEmembership{Senior Member, IEEE,}\\
        Avishek Nag, \IEEEmembership{Senior Member, IEEE,}
        and Deepu John, \IEEEmembership{Senior Member, IEEE,}{\thanks{
This work is partly supported by 1) China Scholarship Council, 2) JEDAI Project under the Horizon 2020 FET Chist-Era Program 3) Science Foundation Ireland under Grant number 18/CRT/6183 and 4) Microelectronic Circuit Centre Ireland.

Guoxin Wang, Qingyuan Wang, Avishek Nag, and Deepu John are with the School of Electrical and Electronic Engineering, University College Dublin, Dublin 4, Ireland (e-mail: \{guoxin.wang, qingyuan.wang\}@ucdconnect.ie; \{avishek.nag, deepu.john\}@ucd.ie;) and Ganesh Neelakanta Iyer is with Department of Computer Science, National University of Singapore (e-mail: gni@nus.edu.sg).
}}}

\maketitle

\begin{abstract}
Unsupervised learning methods have become increasingly important in deep learning due to their demonstrated large utilization of datasets and higher accuracy in computer vision and natural language processing tasks. There is a growing trend to extend unsupervised learning methods to other domains, which helps to utilize a large amount of unlabelled data. This paper proposes an unsupervised pre-training technique based on masked autoencoder (MAE) for electrocardiogram (ECG) signals. In addition, we propose a task-specific fine-tuning to form a complete framework for ECG analysis. The framework is high-level, universal, and not individually adapted to specific model architectures or tasks. Experiments are conducted using various model architectures and large-scale datasets, resulting in an accuracy of 94.39\% on the MITDB dataset for ECG arrhythmia classification task. The result shows a better performance for the classification of previously unseen data for the proposed approach compared to fully supervised methods. 
% $\footnote{Code available at {\color{red}https://}}$
\end{abstract}

\begin{IEEEkeywords}
Masked Autoencoder, Unsupervised Learning, Big Data, Electrocardiogram
\end{IEEEkeywords}

\section{Introduction}
Electrocardiogram (ECG) analysis is crucial in diagnosing heart disease and related biomedical applications \cite{Luz2016Apr}. 
Typically, characteristic features of ECG such as time intervals, amplitude, and statistical parameters are extracted, and traditional machine learning methods are used to analyze ECG based on these features \cite{Hong2017Sep}.
%Typically, ECG characteristics such as interval, frequency, and statistical information, among others, are used at the beginning of a session to extract features. Machine learning methods are usually used to classify these extracted features \cite{Hong2017Sep}. 
More recently, deep learning methods have demonstrated improved performance and effectiveness in biomedical signal analysis \cite{10198098, 10078241}.
%Deep learning methods have emerged as promising approaches in this field to enhance performance and have shown effectiveness in recent state-of-the-art \cite{10198098, 10078241}. 
These methods apply supervised learning, and results are often improved by designing better model architectures. One of the benefits of deep learning is that it facilitates the extraction of high-dimensional features from the signal without the need for complex manual pre-processing. In \cite{Takalo-Mattila2018Aug}, Takalo-Mattila et al. built an automatic ECG classification system using Convolutional-Neural-Network (CNN)-based feature extraction. 
A multilayer perceptron (MLP) is used to classify ECG beats. This framework achieves an accuracy of $89.9\%$ when tested with $49712$ samples. Li et al. \cite{Li2022Feb} presented an arrhythmia classification method that extracted ECG features by Residual Neural Network (ResNet) and enhanced it by overlapping segmentation method. They reported an accuracy of $88.9\%$ over $7942$ subjects.

However, traditional supervised learning methods heavily rely on annotated labels from a single dataset; model training tends to overfit because the dataset is too small and has possible label errors, limiting the resulting models' generalization capability. In addition, many methods claiming excellent results are performed on intra-patient tasks, a division that introduces the same record into the training and testing sets, and, therefore, the high performance that cannot be obtained in real scenarios. Moreover, these methods often employ engineering techniques that may introduce data leakage or biases, raising concerns about the credibility of the reported results. In contrast, unsupervised learning methods offer a compelling alternative as they do not require labelled data, enabling the utilization of larger datasets while reducing errors associated with manual annotation. Masked autoencoder (MAE) is an effective, simple, unsupervised representation learning strategy proven in computer vision tasks \cite{he2022masked}, which could be extended to other research areas \cite{9332012, 9401741}.

This paper introduces a novel unsupervised pre-training technique based on the MAE for ECG-related applications and leveraging data augmentation techniques to improve performance. A task-specific fine-tuning is proposed for downstream applications. The complete framework is presented systematically, encompassing all essential aspects ranging from training to testing. To assess its efficacy, we choose cardiac arrhythmia classification as a case task, and the framework's performance is thoroughly evaluated through simulations, providing valuable information about its capabilities and potential clinical utility.

\begin{figure*}[htbp]
    \centering
    \includegraphics[width=0.9\linewidth]{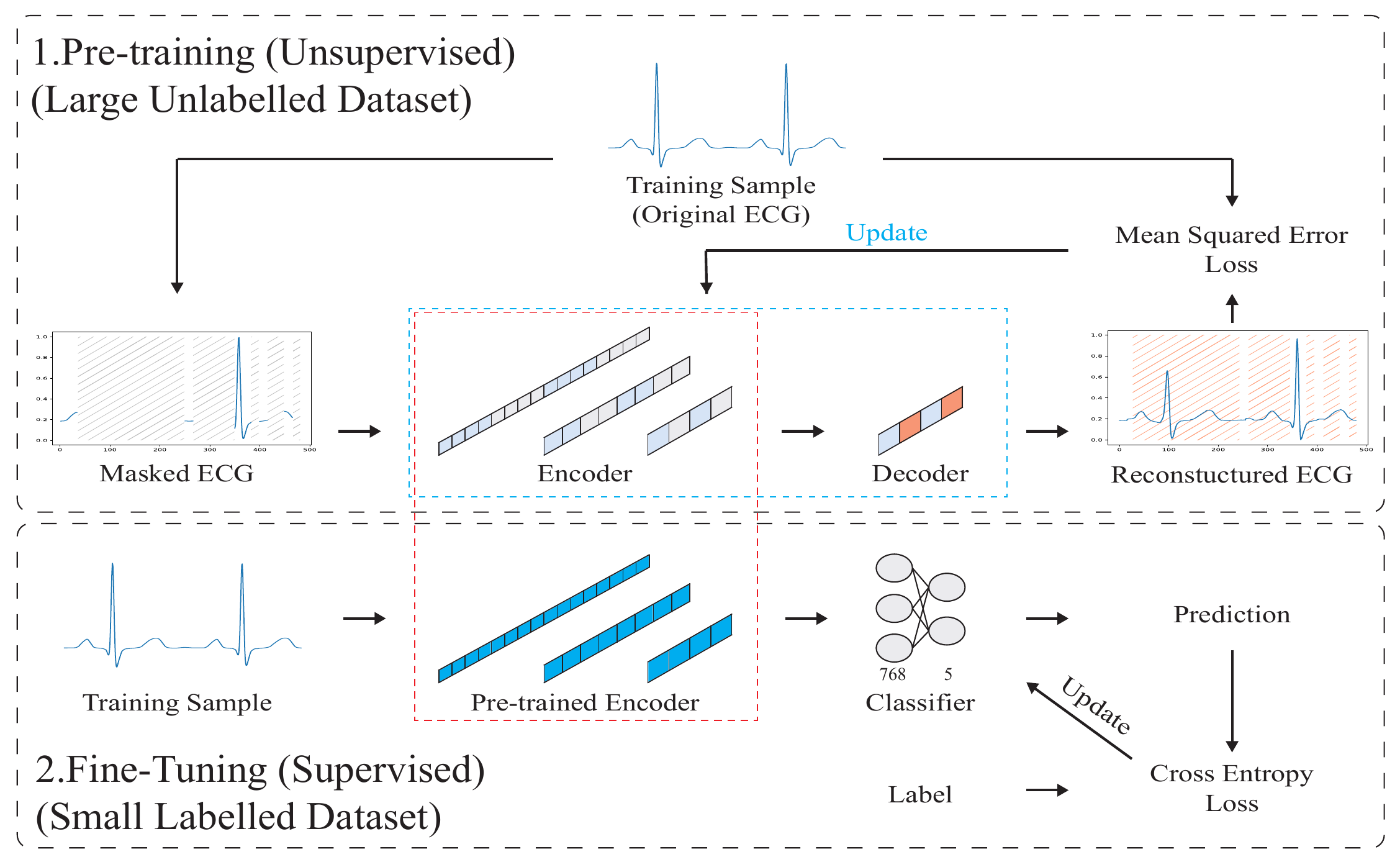}
    \caption{Overview of the MAE-based unsupervised pre-training technique with downstream fine-tuning}
    \label{fig_1_overview}
\end{figure*}

The contributions of this research are as follows:

\begin{itemize}
    \item Unsupervised Pre-training and Task-specific Fine-tuning for ECG: This study presents a novel framework based on MAE for ECG signal analysis. The proposed framework achieves an accuracy of 94.39\% on classifying cardiac arrhythmias in previously unseen data. Using unsupervised learning techniques, the framework overcomes the limitations of traditional supervised methods, which require extensive manual labelling of ECG records.
    
    \item Using Larger-Scale datasets: Unlike conventional approaches that rely on labour-intensive labelling of individual ECG records, the proposed pre-training reduces the need for independent annotations. This feature facilitates more accessible and more efficient model training, enabling the utilization of large-scale datasets without the requirement of extensive manual labelling. This significantly contributes to improved scalability and potential for real-world implementations.
\end{itemize}

The remainder of this paper is organized as follows: Section II details the proposed unsupervised learning-based technique and fine-tuning. The experimental results are presented in Section III, and Section IV provides conclusions and directions for future work.

\section{Method}
The complete framework comprises two main parts: pre-training and fine-tuning. In the pre-training phase, we develop a training strategy for electrocardiogram signals based on the MAE. We then train the base model using a sizeable unlabelled dataset. In the fine-tuning phase, we freeze the base model and train the classifier using a small labelled dataset for the specific task. An overview of the framework is presented in \textit{Figure \ref{fig_1_overview}}.

\subsection{Pre-train}
The predominant approach to classifying cardiac arrhythmias involves supervised training using a limited dataset. However, this method has a potential drawback in that the trained model can become over-fitted, resulting in satisfactory performance on the training dataset but poor generalizability to other datasets \cite{Ying2019Feb}. To alleviate this issue, we propose using an unsupervised learning method. To this end, MAE-based training has been devised, described in greater detail below.

\begin{figure}[ht]
    \centering
    \subfloat[Original]{
        \label{fig_2_1_mae_ori}
        \includegraphics[width=0.4\linewidth]{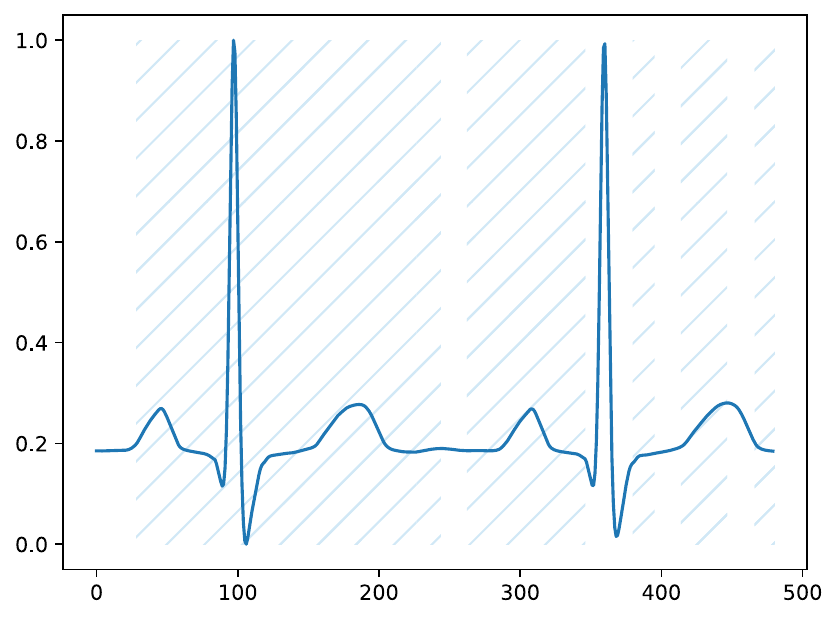}}\\
    \subfloat[Masked]{
        \label{fig_2_2_mae_masked}
        \includegraphics[width=0.4\linewidth]{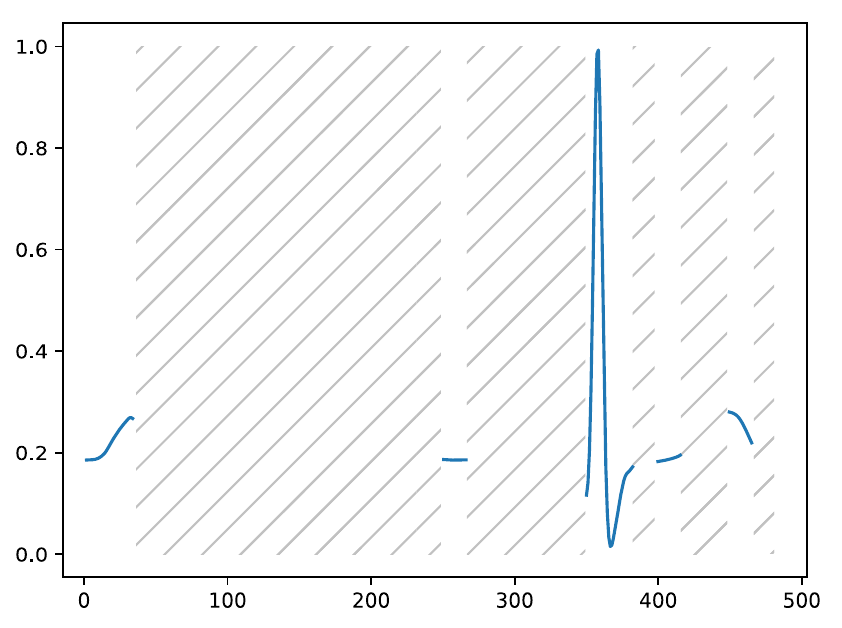}}
    \subfloat[Reconstructed]{
        \label{fig_2_3_mae_resconst}
        \includegraphics[width=0.4\linewidth]{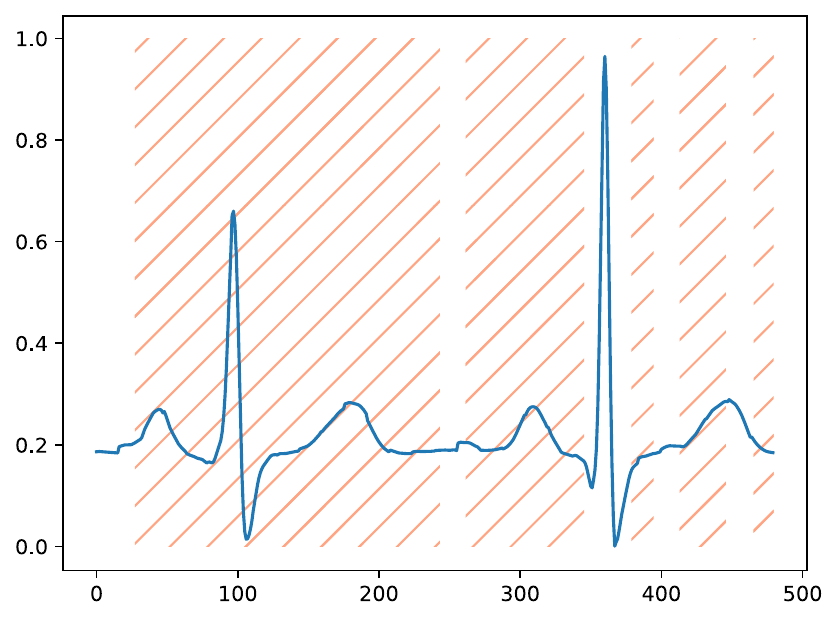}}
    \caption{An example of MAE, which attempts to reconstruct the original signal with limited information from the masked signal.}
    \label{fig_2_mae}
\end{figure}

The MAE-based solution is conceptually simple in that it removes a portion of the data and learns to predict what was removed. Its effectiveness has also been proven in computer vision and natural language processing. Specifically, this encoder-decoder structure operates by dividing input data into patches, with the encoder only processing a visible subset of patches, as depicted in \textit{Figure \ref{fig_2_2_mae_masked}}. Subsequently, the decoder reconstructs the input with incomplete information. Although the reconstructed output may not be perfect, this approach helps the model better comprehend the input. Once trained, the decoder can be removed, and the encoder can serve as a practical feature extractor in other related tasks. An instance of the MAE applied to ECG signals is shown in \textit{Figure \ref{fig_2_mae}}.  MAE is beneficial for using large unlabelled datasets and can be advantageous for various downstream tasks, particularly for ECG classification with limited annotated data. In this paper, we utilize the one-dimensional version of the MAE.

\begin{figure}[ht]
    \centering
    \includegraphics[width=\linewidth]{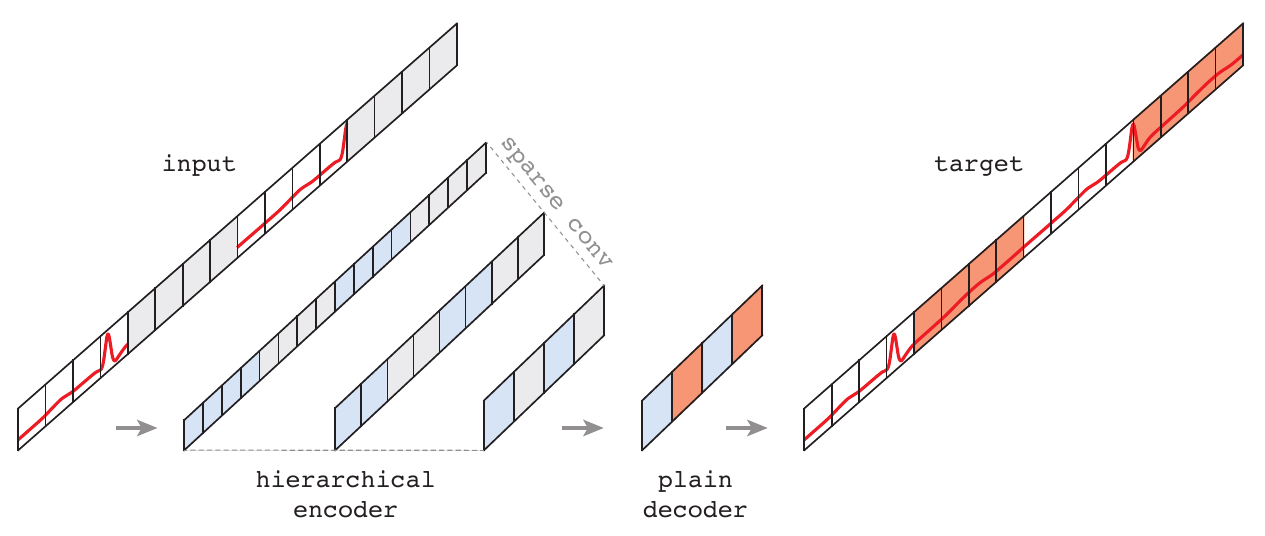}
    \caption{ConvNeXtV2-1D for ECG applications}
    \label{fig_3_model}
\end{figure}

This paper utilizes ConvNeXtV2 \cite{woo2023convnext} to implement a fully convolutional MAE (FCMAE), which is state-of-the-art in the image classification task. This method employs a non-symmetric encoder-decoder design and sparse convolution to reduce computational burden during the pre-training phase. The original ConvNeXtV2 model was designed for images, whereas our implementation ConvNeXtV2-1D includes the necessary modifications to accommodate the 1D ECG signal. The architecture is illustrated in \textit{Figure \ref{fig_3_model}}.

\subsection{Fine-tune}

\subsubsection{Data Augmentation}
Data augmentation enhances the model's generalization performance in fine-tuning. Various augmented methods include mixup as described in \cite{Zhang2017Oct} and additional white noise. By combining these methods, we ensured that the fundamental characteristics of the original data, such as the relative positions of fiducial points and peak intervals, remained unchanged.

\subsubsection{Modelling}
We extended the pre-trained model into our framework by adding an MLP head as a classifier for detecting arrhythmia. We used well-established forward and backward propagation techniques for supervised learning during this phase. However, it is essential to note that the pre-trained model remained frozen throughout this process, ensuring that the previously learned features were retained without alteration. Therefore, only the newly added classifier underwent training, allowing it to specialize in accurately identifying arrhythmia patterns.

\section{Evaluation and Results}
\subsection{Datasets}
This research used multiple datasets, each serving specific purposes. The first dataset used for pre-training the model was the PhysioNet / Computing in Cardiology Challenge 2021 (CINC2021) introduced by Alday et al. \cite{Alday2020Dec}. This large-scale database assembled nine databases with $131,149$ unlabelled 12-lead ECG records for over 1000 hours.

For the subsequent stages of fine-tuning and testing, we employed the MIT-BIH Arrhythmia Database (MITDB) curated by Moody and Mark \cite{moody1992mit}. This specific database consists of 48 records derived from 47 individual subjects. We also used the St Petersburg INCART 12-lead Arrhythmia Database (INCARTDB) \cite{tihonenko2007st} for fine-tuning. The INCARTDB database includes 75 records from 32 subjects.

To partition the cardiac rhythm data extracted from the MIT-BIH Arrhythmia Database (MITDB), we adopt a methodology similar to the approach employed by \cite{1306572}. The dataset denoted as DS1 is used for fine-tuning, while DS2 serves as the designated testing dataset. The heartbeats of DS1 and DS2 come from different individuals. Such division protocol is called in the literature inter-patient paradigm \cite{6036156}. Furthermore, DS1 is divided into a ratio of 9:1, where 90\% of the data are allocated for training purposes, while the remaining 10\% are allocated for validation. For INCARTDB, we use the whole dataset for fine-tuning.

The MITDB and INCARTDB database contains labelled annotations for four classes: N-type (normal beat), SVEB-type (atrial premature beat), VEB-type (premature ventricular contraction), F-type (fusion of ventricular and normal beat), and Q-type (unknown beat). Table \ref{tab_1_classes} summarises the data distribution for these classes.

\begin{table}[htbp]
\centering
\captionsetup{justification=centering}
\caption{data distribution for labelled data}
\label{tab_1_classes}
\resizebox{\columnwidth}{!}{%
\begin{tabular}{cccccc}
\hline
\multirow{2}{*}{}          & \multirow{2}{*}{N}      & \multirow{2}{*}{SVEB} & \multirow{2}{*}{VEB}   & \multirow{2}{*}{F}   & \multirow{2}{*}{Q} \\
                           &                         &                       &                        &                      &                    \\ \hline
\multirow{2}{*}{INCARTDB}  & \multirow{2}{*}{153563} & \multirow{2}{*}{1958} & \multirow{2}{*}{20000} & \multirow{2}{*}{219} & \multirow{2}{*}{6} \\
                           &                         &                       &                        &                      &                    \\
\multirow{2}{*}{MITDB-DS1 (Training)}   & \multirow{2}{*}{45298} & \multirow{2}{*}{908} & \multirow{2}{*}{3597} & \multirow{2}{*}{393} & \multirow{2}{*}{9} \\
                           &                         &                       &                        &                      &                    \\
\multirow{2}{*}{MITDB-DS1 (Validation)} & \multirow{2}{*}{2392}  & \multirow{2}{*}{38}  & \multirow{2}{*}{190}  & \multirow{2}{*}{21}  & \multirow{2}{*}{1} \\
                           &                         &                       &                        &                      &                    \\
\multirow{2}{*}{MITDB-DS2} & \multirow{2}{*}{44225}  & \multirow{2}{*}{1837} & \multirow{2}{*}{3219}  & \multirow{2}{*}{388} & \multirow{2}{*}{7} \\
                           &                         &                       &                        &                      &                    \\ \hline
\end{tabular}%
}
\end{table}

In the pre-processing step of ECG signals, segmenting them into shorter pieces using fiducial points is standard practice. Initially, we re-sampled the ECG signals at a frequency of 360 Hz. Then, we detect all the 'R' peaks in the ECG signals for the unlabelled dataset. For labelled datasets, we use annotations, including peak position and diagnosis information. Next, we extract a segment of 480 sample points for each 'R' peak. This segment encompasses 360 points to the left and 120 points to the right of the 'R' peak. To ensure consistency across instances, we normalize each segment to a range between 0 and 1 in the final.

\begin{table*}[htbp]
\renewcommand\thetable{IV}
\centering
\captionsetup{justification=centering}
\caption{Comparison of the proposed system with other published works}
\label{system_comparision}
\resizebox{\textwidth}{!}{%
\begin{tabular}{cccccc}
\hline
\multirow{2}{*}{Work} &
  \multirow{2}{*}{Feature Exactor} &
  \multirow{2}{*}{Classifier} &
  \multirow{2}{*}{Training Set} &
  \multirow{2}{*}{Testing Set} &
  \multirow{2}{*}{Accuracy} \\
 &
   &
   &
   &
   &
   \\ \hline
\multirow{2}{*}{Takalo-Mattila et al. \cite{Takalo-Mattila2018Aug}} &
  \multirow{2}{*}{CNN} &
  \multirow{2}{*}{MLP} &
  \multirow{2}{*}{MITDB-DS1} &
  \multirow{2}{*}{MITDB-DS2} &
  \multirow{2}{*}{89.9\%} \\
 &
   &
   &
   &
   &
   \\
\multirow{2}{*}{Li et al. \cite{Li2022Feb}} &
  \multirow{2}{*}{ResNet} &
  \multirow{2}{*}{MLP} &
  \multirow{2}{*}{MITDB-DS1} &
  \multirow{2}{*}{MITDB-DS2} &
  \multirow{2}{*}{88.9\%} \\
 &
   &
   &
   &
   &
   \\
\multirow{3}{*}{Sellami et al. \cite{Sellami2019May}} &
  \multirow{3}{*}{CNN} &
  \multirow{3}{*}{MLP} &
  \multirow{3}{*}{MITDB-DS1} &
  \multirow{3}{*}{MITDB-DS2} &
  \multirow{3}{*}{88.3\%} \\
 &
   &
   &
   &
   &
   \\
 &
   &
   &
   &
   &
   \\
\multirow{2}{*}{Lin et al. \cite{Lin2014May}} &
  \multirow{2}{*}{Morphological Features} &
  \multirow{2}{*}{Linear Discriminant} &
  \multirow{2}{*}{MITDB-DS1} &
  \multirow{2}{*}{MITDB-DS2} &
  \multirow{2}{*}{91.6\%} \\
 &
   &
   &
   &
   &
   \\
\multirow{2}{*}{Asl et al. \cite{Asl2008Sep}} &
  \multirow{2}{*}{Generalized Discriminant Analysis} &
  \multirow{2}{*}{SVM} &
  \multirow{2}{*}{MITDB-DS1 + MITDB-DS2} &
  \multirow{2}{*}{MITDB-DS1 + MITDB-DS2} &
  \multirow{2}{*}{100\%} \\
 &
   &
   &
   &
   &
   \\
\multirow{2}{*}{Chen et al. \cite{Chen2014Dec}} &
  \multirow{2}{*}{Fuducial Features} &
  \multirow{2}{*}{SVM + MLP} &
  \multirow{2}{*}{MITDB-DS1 + MITDB-DS2} &
  \multirow{2}{*}{MITDB-DS1 + MITDB-DS2} &
  \multirow{2}{*}{100\%} \\
 &
   &
   &
   &
   &
   \\
\multirow{2}{*}{\textbf{Proposed Method}} &
  \multirow{2}{*}{\textbf{Unsupervised Pre-training}} &
  \multirow{2}{*}{\textbf{MLP}} &
  \multirow{2}{*}{\textbf{MITDB-DS1 + INCARTDB}} &
  \multirow{2}{*}{\textbf{MITDB-DS2}} &
  \multirow{2}{*}{\textbf{94.39\%}} \\
 &
   &
   &
   &
   &
   \\ \hline
\end{tabular}%
}
\end{table*}

\subsection{Architecture}
Our proposed framework is trained at a high level, and we evaluate it using different architectures tailored explicitly to the model. To leverage the simplicity of implementation, we utilize multiple variations of ConvNeXtV2-1D in the MAE-based training and the supervised training baseline for comparison and benchmarking purposes. We follow the same configurations of the stage, block ($B$), and channel ($C$) settings in \cite{woo2023convnext}.

\begin{itemize}
    \item ConvNeXtV2-1D-Atto: $C=40$, $B=(2, 2, 6, 2)$
    \item ConvNeXtV2-1D-Tiny: $C=96$, $B=(3, 3, 9, 3)$
    \item ConvNeXtV2-1D-Base: $C=192$, $B=(3, 3, 27, 3)$
\end{itemize}

\subsection{Experiment Setup}
The pre-trained model uses Stochastic Gradient Descent (SGD) with a batch size of 512 for 500 epochs.  Adam optimizer is employed with a batch size of 1024 for 100 epochs for fine-tuning, initializing the learning rate to 0.0003. The learning rate is gradually reduced using cosine annealing.

To establish a benchmark, we conduct complete supervised training using the same parameters as the fine-tuning process.

\subsection{Evaluation}

MITDB-DS2 is the test set to calculate global performance for different methods. Accuracy is the main critical metric for the classification task. \textit{Table \ref{tab_2_f1}} shows detailed results of different model architectures and training strategies. The results show that the proposed method achieved a higher accuracy than traditional supervised training, which is 90.83\% for ConvNeXtV2-1D-Atto, 91.30\% for ConvNeXtV2-1D-Tiny and 91.34\% for ConvNeXtV2-1D-Base.

\begin{table}[htbp]
\renewcommand\thetable{II}
\centering
\captionsetup{justification=centering}
\caption{accuracy with different model architectures and training strategies}
\label{tab_2_f1}
\resizebox{\columnwidth}{!}{%
\begin{tabular}{cccc}
\hline
\multirow{2}{*}{}               & \multirow{2}{*}{ConvNeXtV2-1D-Atto} & \multirow{2}{*}{ConvNeXtV2-1D-Tiny} & \multirow{2}{*}{ConvNeXtV2-1D-Base} \\
 &  &  &  \\ \hline
\multirow{2}{*}{\textbf{Proposed Method}} & \multirow{2}{*}{\textbf{90.83\%}}   & \multirow{2}{*}{\textbf{91.30\%}}   & \multirow{2}{*}{\textbf{91.34\%}}   \\
 &  &  &  \\
\multirow{2}{*}{Supervised}     & \multirow{2}{*}{89.48\%}            & \multirow{2}{*}{90.58\%}            & \multirow{2}{*}{88.54\%}            \\
 &  &  &  \\ \hline
\end{tabular}%
}
\end{table}

In addition, \textit{Table \ref{tab_2_f2}} shows that adding different datasets for fine-tuning and supervised learning is helpful for performance improvement. MAE-based training with MITDB-DS1 and INCARTDB fine-tuning achieved higher accuracy among multiple architecture complexity, which is 94.39\% for ConvNeXtV2-1D-Atto, 93.98\% for ConvNeXtV2-1D-Tiny and 93.89\% for ConvNeXtV2-1D-Base.

\begin{table}[htbp]
\renewcommand\thetable{III}
\centering
\captionsetup{justification=centering}
\caption{accuracy with different fine-tuning datasets}
\label{tab_2_f2}
\resizebox{\columnwidth}{!}{%
\begin{tabular}{cccc}
\hline
\multirow{2}{*}{} &
  \multirow{2}{*}{ConvNeXtV2-1D-Atto} &
  \multirow{2}{*}{ConvNeXtV2-1D-Tiny} &
  \multirow{2}{*}{ConvNeXtV2-1D-Base} \\
 &  &  &  \\ \hline
\multirow{2}{*}{\begin{tabular}[c]{@{}c@{}}Proposed Method\\ (MITDB-DS1)\end{tabular}} &
  \multirow{2}{*}{90.83\%} &
  \multirow{2}{*}{91.30\%} &
  \multirow{2}{*}{91.34\%} \\
 &  &  &  \\
\multirow{2}{*}{\textbf{\begin{tabular}[c]{@{}c@{}}Proposed Method\\ (MITDB-DS1, INCARTDB)\end{tabular}}} &
  \multirow{2}{*}{\textbf{94.39\%}} &
  \multirow{2}{*}{\textbf{93.98\%}} &
  \multirow{2}{*}{\textbf{93.89\%}} \\
 &  &  &  \\ \hline
\end{tabular}%
}
\end{table}

\textit{Table \ref{system_comparision}} compares the proposed framework and the current, reliable state of the art. \cite{Takalo-Mattila2018Aug, Li2022Feb, Sellami2019May}, who have reported performance on the MITDB-DS2 with supervised deep learning methods, which is less than ours. Furthermore, their system used a single dataset for training and did not take full advantage of the available ECG dataset, resulting in low accuracy. Furthermore, \cite{Lin2014May} used methods based on machine learning and required a lot of manual handling but also reported low precision. \cite{Asl2008Sep, Chen2014Dec} have reported an accuracy of 100\%, but they actually conducted the intra-patient test, where heartbeats of the same records probably appear in training and the testing dataset. However, in a realistic scenario, a fully automatic method will find patients' heartbeats different from those they used to learn in the training phase; hence, the high accuracy they report is questionable. Compared with these frameworks, our proposed framework achieves higher accuracy within the same task, fully uses existing datasets, and meets practical needs.

\section*{IV Conclusion}
Our study introduced a new MAE-based cardiac arrhythmia classification system. The system uses unsupervised learning to learn generic ECG information and classify arrhythmia after fine-tuning. Experiments show that the proposed approach improves performance compared to traditional methods. Future work includes using different unsupervised learning approaches, more architectures, larger datasets, model compression, embedded system deployment, and transfer learning exploration.
\clearpage
\bibliographystyle{unsrt}
\bibliography{reference}

\end{document}